\documentclass{article}

     \PassOptionsToPackage{numbers, compress}{natbib}

\usepackage{amsmath}
\usepackage{amssymb}
\usepackage{algorithm}
\usepackage{algpseudocode}
\algnewcommand{\algorithmicforeach}{\textbf{for each}}
\algdef{SE}[FOR]{ForEach}{EndForEach}[1]
  {\algorithmicforeach\ #1\ \algorithmicdo}
  {\algorithmicend\ \algorithmicforeach}

\usepackage{float}

\usepackage[preprint]{neurips_2021}
\usepackage{graphicx}
\usepackage[utf8]{inputenc} 
\usepackage[T1]{fontenc}    
\usepackage{hyperref}       
\usepackage{url}            
\usepackage{booktabs}       
\usepackage{amsfonts}       
\usepackage{nicefrac}       
\usepackage{microtype}      
\usepackage{xcolor}         
\usepackage{multicol}
\usepackage{bm}
\usepackage{mathrsfs}
\usepackage{soul}

\title{Making CNNs Interpretable by Building Dynamic Sequential Decision Forests with Top-down Hierarchy Learning}

\author{Yilin Wang$^1$, Shaozuo Yu$^2$, Xiaokang Yang$^1$, Wei Shen$^1$\\
$^1$ MoE Key Lab of Artificial Intelligence, AI Institute, Shanghai Jiao Tong University\\
$^2$ Department of Computer Science and Technology, Tongji University\\
{\tt\footnotesize
\{wangyilin210210,yuyuanbo1957\}@gmail.com \{xkyang,wei.shen\}@sjtu.edu.cn}}

\begin{document}

\maketitle

\begin{abstract}

  In this paper, we propose a generic model transfer scheme to make Convlutional Neural Networks (CNNs) interpretable, while maintaining their high classification accuracy. We achieve this by building a differentiable decision forest on top of CNNs, which enjoys two characteristics: 1) During training, the tree hierarchies of the forest are learned in a top-down manner under the guidance from the category semantics embedded in the pre-trained CNN weights; 2) During inference, a single decision tree is dynamically selected from the forest for each input sample, enabling the transferred model to make sequential decisions corresponding to the attributes shared by semantically-similar categories, rather than directly performing flat classification. We name the transferred model deep Dynamic Sequential Decision Forest (dDSDF). Experimental results show that dDSDF not only achieves higher classification accuracy than its conuterpart, \emph{i.e.}, the original CNN, but has much better interpretability, as qualitatively it has plausible hierarchies and quantitatively it leads to more precise saliency maps.
\end{abstract}

\section{Introduction}

In recent years, Convolutional Neural Networks (CNNs)~\cite{Ref:LeCunBDHHHJ89,Ref:Krizhevsky12,Ref:SimonyanZ14a} have become the dominant models for vision recognition tasks, such as image classification, thanks to their powerful representation learning ability and outstanding performance. However, they are criticized for the lack of interpretability~\cite{Ref:YuilleL21}. This drawback inevitably increases unpredictable risks when applying CNNs to real-world computer vision applications concerned with model reliability, such as medical image diagnosis and autonomous driving. The reason why CNNs are known as black box models is it is difficult for humans to understand their principle to make a prediction. For example, humans can naturally perform hierarchical classification with semantically-plausible sequential decisions, e.g., canine? $\rightarrow$ blue eyes? $\rightarrow$ Huskie, but CNNs only perform flat classification, impeding understanding their decisions.

On the contrary, decision trees~\cite{Ref:BreimanFOS84}, which make sequential decisions during inference, were among the most popular machine learning models for various vision recognition tasks, given their simplicity and interpretability~\cite{Ref:Gareth15,Ref:WuKQGYMMNLYZSHS08}. In light of the complementary properties of decision trees to CNNs, a large amount of efforts have been made to combine these two worlds, with the purpose to build a better model which is able to provide both high performance and good interpretability. However, these attempts barely live up to expectation. They suffer from 1) sacrifice of interpretability due to pursuing high performance~\cite{Ref:KontschiederFCB15,Ref:ShenGWZWY21}; 2) performance degradation due to imposing interpretability~\cite{Ref:TannoAACN19,Ref:AlanizA19}; 3) lack of generalizability, \emph{i.e.}, they are only designed for customized CNNs~\cite{Ref:ZhangYMW19}.

To address these issues, we propose a novel scheme to combine CNNs and decision trees, which is a generic model transfer scheme, with the ability to make any CNNs interpretable while maintaining their high classification accuracy. Given a pre-trained CNN, we transfer it to an interpretable model by building a differentiable decision forest (tree ensemble)~\cite{Ref:KontschiederFCB15} on top of it: Each tree split node is connected to a neuron of the last fully-connected layer of the CNN, and thus the decision made at each split node is determined by the output of its corresponding neuron. Intuitively, this model transfer scheme can guarantee high model accuracy, thanks to joint tree ensemble learning and representation learning of the CNN. To make the transferred model interpretable, we design two mechanisms for forest building: 1) A top-down hierarchy learning mechanism, which imposes interpretable semantics to the sequential decisions along the tree paths from the root to leaf nodes. Concretely, we design a criterion to form the correspondence between the split nodes and the neurons under the guidance from the category semantics embedded in the pre-trained weights of the CNN~\cite{Ref:QiaoLSY18,Ref:QiBL18}. This criterion leads to a hierarchy which implicitly clusters semantically-similar categories in a top-down manner, so that they can share the same decision path, from which some semantically-plausible attributes can be extracted to explain each decisions. 2) A dynamic tree ensemble thinning mechanism, which selects one single most representative tree for each input sample during inference, so that the ensemble can be interpreted~\cite{Ref:Hernandez-LobatoMS09,Ref:VidalS20}. We name the transferred model deep Dynamic Sequential Decision Forest (dDSDF), as it dynamically queries a single tree predictor from the forest and makes sequential semantically-plausible decisions on top of deep networks. We further propose a decision-tree-based Class Activation Map (CAM)~\cite{Ref:ZhouKLOT16,Ref:SelvarajuCDVPB20} approach (as only one tree is selected for an input sample during inference), and show that dDSDF can generate more precise saliency maps than its conuterpart, \emph{i.e.}, the original CNN, to explian its prediction.

Experimental results on several benchmark datasets, such as Cifar~\cite{Ref:Krizhevsky09}, tinyImageNet~\cite{Ref:Tiny} and ImageNet~\cite{Ref:Russakovsky15}, verify the benefits of dDSDF: 1) it can achieve higher classification accuracy than its conuterpart, \emph{i.e.}, the original CNN; 2) it has much better interpretability, since qualitatively it has a semantically-plausible hierarchy and quantitatively it leads to more precise saliency maps.

\section{Related Work}
In recent years, a large amount of efforts have been made to combine deep networks and decision trees for either higher performance or better interpretability.
\subsection{Combination for Higher Performance}

\paragraph{Building deep networks with tree-like architectures.} By this strategy, a data sample only visit a fraction of neurons in networks. Ioannou~\emph{et al.}~\cite{Ref:Ioannou16} proposed Conditional Networks, in which data routers are introduced, represented as perceptrons, to send incoming data to a selected sub-branch. Tanno~\emph{et al.}~\cite{Ref:TannoAACN19} proposed Adaptive Neural Trees, which additionally learns tree topologies by greedily searching three tree growing choices: splitting, keeping and deepening. Roy~\emph{et al.}~\cite{Ref:RoyPR20} proposed a CNN with tree structure, which is built by growing the CNN in a tree-like fashion, to deal with data with unseen classes. Murthy~\emph{et al.}~\cite{Ref:MurthySCMC16} presented a tree-like structured network model driven by the data. Starting from the root network node, this tree-like structured network model automatically builds a network that splits the hard examples into disjoint clusters of classes which would be handled by the subsequent expert networks. Xiong~\emph{et al.}~\cite{Ref:Xiong15} proposed a conditional Convolutional Neural Network (c-CNN) to handle multimodal face recognition. In c-CNN, face samples of different modalities were passed along with modality-specific routes, gradually separated layer by layer and finally passed into different leaf nodes. This combination strategy suffers from enormous additional parameters compared to normal neural networks, since the data routers are often represented as another deep routing network. Besides, such models always require elaborate network designs for specific tasks, making them difficult to reuse and transfer.

\paragraph{Building decision trees on top of deep networks.} This strategy defines the split functions of the tree according to the output of networks or neurons, which can be directly benefited from existing sophisticated deep networks~\cite{Ref:SimonyanZ14a,Ref:HeZRS16}.  Bul\`{o} and Kontschieder~\cite{Ref:BuloK14} presented randomized Multi-Layer Perceptrons (rMLP) as new split functions which are capable of learning non-linear, data-specific representations and taking advantage of them by finding optimal predictions for the emerging child nodes. By introducing rMLP, data representation and discriminative learning within randomized decision trees can be jointly tackled. However, representations were learned only locally at split node level and independently among split nodes. Kontschieder~\emph{et al.}~\cite{Ref:KontschiederFCB15} proposed deep Neural Decision Forests (dNDFs), which connect each split node to a neuron in a fully-connected (FC) layer of a deep network. A probabilistic split function is defined at each split node according to the output value of the corresponding neuron and a global loss function is defined on a tree. This ensures that the split node parameters and leaf node predictions can be learned jointly with the deep network. This combination strategy has a good property: It can jointly optimize network parameters, data space partition at split nodes and data distribution abstraction at leaf nodes. Thereby, a lot of works followed this line. Roy and Todorovic~\cite{Ref:Roy16} represented each split function by a small CNN, and used this tree-based CNN for depth estimation. Chen~\emph{et al.}~\cite{Ref:ChenHTWC16} extend dNDFs to deal with domain adaptation problems. Zhu~\emph{et al.}~\cite{Ref:ZhuSMYRZ17} proposed Deep Embedding Forest on the basis of dNDFs for deep text feature mining. Shen~\emph{et al.}~\cite{Ref:Shen17,Ref:Shen18,Ref:ShenGWZWY21} extended dNDFs to perform label distribution learning and regression by proposing Label Distribution Learning Forest (LDLF) and deep Regression Forest (dRF), respectively, and verified the effectiveness of LDLF and dRF on age estimation. Pan~\emph{et al.}~\cite{Ref:PanA0X20} further introduced self-paced learning into dRF. Although these dNDF-based models achieved excellent prediction performance, the interpretability of decision trees was sacrificed. The reason is two-fold: 1) The correspondences between split nodes and neurons in dNDFs are randomly assigned, resulting in tree hierarchies without category semantics and unexplainable decisions during inference; 2) The forest sacrifices the intrinsic interpretability present in decision trees, since following the decision paths of the ensemble of trees becomes intractable. The proposed dDSDF addressed the first issue by explicitly learn the correspondence between split nodes and neurons, leading to sequential decisions corresponding to the attributes shared by semantically-similar categories; And it addressed the second issue by introducing the dynamic tree ensemble thinning mechanism, which selects one single most representative decision tree during inference for each input sample.

\subsection{Combination for Better Interpretability}
As a well-recognized interpretable model, leveraging decision trees to explain neural networks is intuitive. Frosst and Hinton~\cite{Ref:FrosstH17} achieved this by distilling the knowledge acquired by a deep network into a soft decision tree. Hehn~\emph{et al.}~\cite{Ref:HehnKH20} introduced a greedy tree structure construction scheme to build unbalance DNDFs with data-specific structures for better interpretability. However, the model obtained by this scheme only works on small-scale datasets, such as MNIST~\cite{Ref:Lecun98}. The scalability of the scheme is questionable, due to some GPU-unfriendly operations in tree structure construction. Zhang~\emph{et al.}~\cite{Ref:ZhangYMW19} enforced a locality constraint onto CNN filters, so that each channel of the CNN is endowed with a specific part of an image. They then constructed a decision tree on top of the CNN and explained the principle of the CNN's decision making by traversing the decision tree from top to bottom, forming a path from common parts of general categories to unique parts of a small number of samples. However, this model transfer scheme is only applicable to the customized CNN they designed, which limits its usage on general CNNs. Wan~\emph{et al.}~\cite{Ref:Wan21} proposed Neural-Backed Decision Tree (NBDT), which transfers a CNN to a interpretable model by building a decision tree on top of the CNN. The decision tree is formed by performing agglomerative hierarchical clustering based on the category similarities embedded in pre-trained CNN weights, and can be fine-tuned with the CNN to enjoy the benefit of high performance. However, this clustering-based bottom-up tree construction scheme solely relies on pre-trained CNN weights, thus might not lead to human-understood hierarchy on a large dataset, \emph{e.g.}, ImageNet~\cite{Ref:Russakovsky15}, without the pre-defined WordNet~\cite{Ref:Miller95}. Unlike NBDT, the each tree in the proposed dDSDF is constructed in a top-down manner, enable us to consider the consistency between the category similarities and category-level statistical routing similarities from top to down in the constructed hierarchy.

\section{Methodology}
\subsection{Problem Statement}
Given a CNN model $\mathscr{C}$ with pre-trained weights $\bm{\Theta}$, whose input is an image $\mathbf{x}\in \mathcal{X}$ and output is its category label $y\in\mathcal{Y}=\{1,2,\ldots,C\}$, where $\mathcal{X}$ is image space and $C$ is the number of image categories, our goal is to transfer $\mathscr{C}$ to a interpretable model $\mathscr{I}$, without classification performance degradation. There is no agreement in the literature about the clear definition of model interpretability, but we can simply follow the description from~\cite{Ref:Gareth15} - interpretable models are \emph{desirable to have information providing qualitative understanding of the relationship between joint values of the input variables and the resulting predicted response value}. Decision trees are commonly-accepted interpretable models, since 1) the final decision made by a decision tree can be followed by a decision path, \emph{i.e.}, a sequential decision process; 2) each decision along the path is determined by interpreable feature selection, which is related to a semantically-plausible attribute. The transferred model $\mathscr{F}$ is desired to have these two properties. Next, we introduce how to achieve this by the proposal of deep Dynamic Sequential Decision Forest (dDSDF).

\subsection{Model Overview}
\begin{figure}[H]
  \centering
  \includegraphics[trim=0cm 0cm 0cm 0cm, clip=true, width=0.75\linewidth]{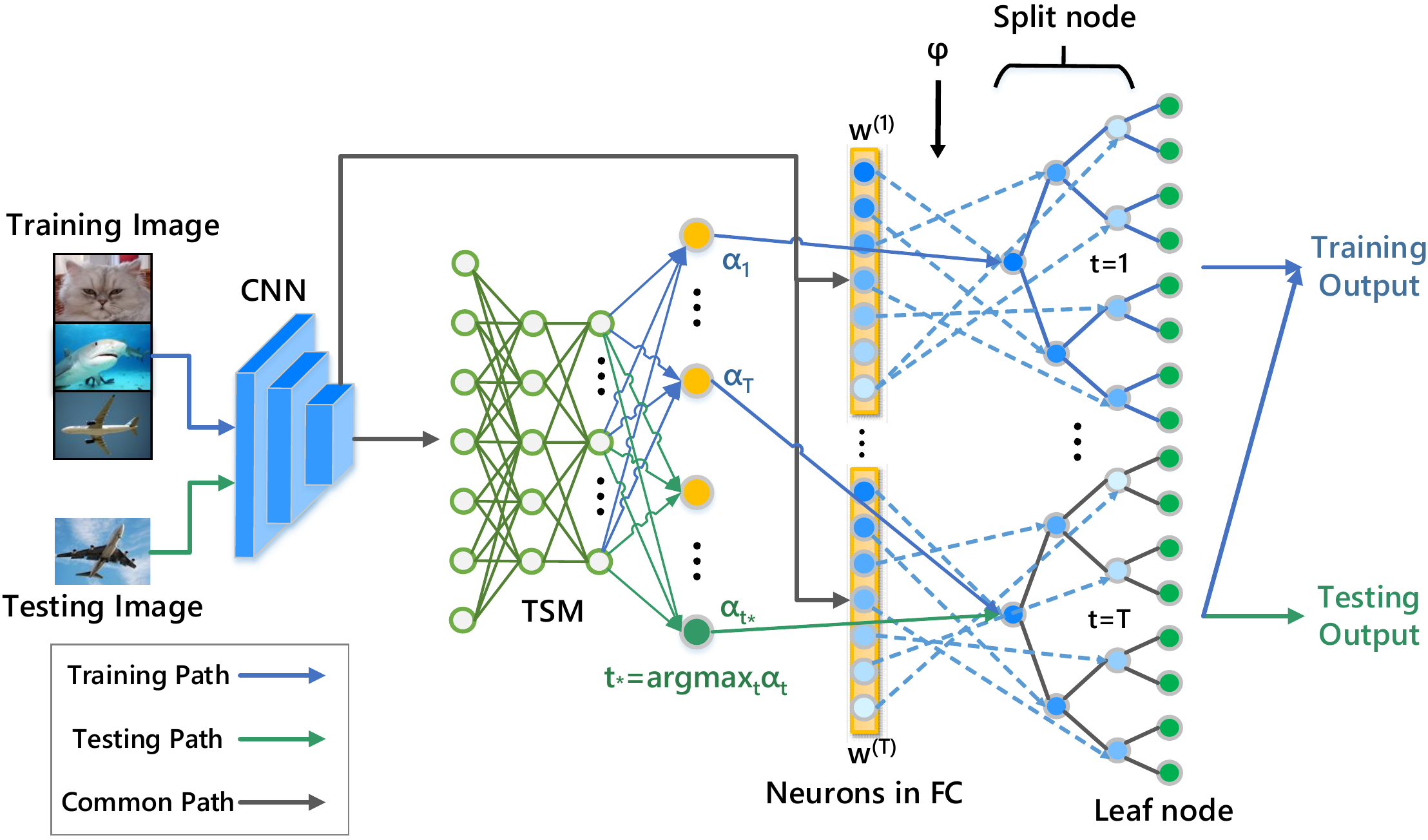} 
  \caption{Deep Dynamic Sequential Decision Forest.}
 \label{fig:DSDF}
\end{figure}

We transfer the CNN model $\mathscr{C}$ to an interpretable model $\mathscr{F}$ by building a differentiable decision forest~\cite{Ref:KontschiederFCB15} (an ensemble of differentiable decision trees) on top of it, as shown in Fig.~\ref{fig:DSDF}: Each tree split node is connected to a neuron of the last fully-connected (FC) layer of the CNN, and thus its decision function is determined by the output of the neuron. With the help of joint CNN and tree ensemble training, the transferred model $\mathscr{F}$ can achieve a high performance. However, the differentiable decision forest sacrifice the intrinsic interpretability present in decision trees, since 1) each tree in the differentiable decision forest is not interpretable, due to random feature selection during tree building; 2) the concurrent use of multiple trees in the ensemble reduces the interpretability~\cite{Ref:VidalS20,Ref:Benard21a}. To respectively address these two issues, we propose 1) a top-down hierarchy learning mechanism, which guides feature selection for each tree split node by the category semantics embedded in the pre-trained CNN weights $\bm{\Theta}$, forming a semantically-plausible sequential decision process for each tree; 2) a dynamic tree selection module (TSM), following the spirit of dynamic tree ensemble thinning~\cite{Ref:Hernandez-LobatoMS09,Ref:VidalS20}, to select one single decision tree for each input sample, so that the ensemble can be interpreted. Next, we first introduce how to make a differentiable decision tree interpretable by top-down hierarchy learning, then describe the scheme to build the Dynamic Sequential Decision Forest $\mathscr{F}$ on top of the CNN $\mathscr{C}$.

\subsection{Building Interpretable Differentiable Decision Tree} \label{sec:tree_building}
\subsubsection{Preliminary} A differentiable decision tree is a full binary tree, thus such a tree with depth $d$ has $N=2^{d-1}-1$ split nodes. To build such a tree on top of the CNN $\mathscr{C}$, we desire a one-to-one correspondence between the split nodes in the tree and the the neurons in the last FC layer of the CNN $\mathscr{C}$, since we expect that the tree structure can represent a plausible category hierarchy. Towards this end, we replace the last FC layer (the layer for $C$-way classification) of the CNN $\mathscr{C}$ by a new one whose dimension is $N$, parameterized by $\mathbf{w}$. Let $\mathcal{M}=\{m_1,m_2,\ldots,m_N\}$ denote the set of neurons in this new FC layer. Then, we build the differentiable decision tree with depth $d$ on the new FC layer: The tree consists of a set of split nodes $\mathcal{N}=\{n_1,n_2,\ldots,n_N\}$ and a set of leaf nodes $\mathcal{L}=\{\ell_1,\ell_2,\ldots,\ell_N\}$. Each leaf node $\ell \in \mathcal{L}$ holds a distribution $\bm{\pi}_\ell=(\pi_{\ell_1},\pi_{\ell_2},\ldots,\pi_{\ell_C})$ over $\mathcal{Y}$. Each split node $n \in \mathcal{N} $ defines a soft decision function $s_n(\mathbf{x}; \bm{\Theta},\mathbf{w}) : \mathcal{X} \rightarrow [0, 1]$ to determine the probability that a sample $\mathbf{x}$ is routed to the left or right sub-tree. Then, the probability of sample $\mathbf{x}$ reaching an arbitrary node $n$ is given by:
\begin{equation} \label{eq:leaf_node_prob}
    \mu(n|\mathbf{x} ; \bm{\Theta},\mathbf{w})=\prod_{n \in \mathcal{N}} s_{n}(\mathbf{x} ; \bm{\Theta},\mathbf{w})^{\mathbf{1}\left(\ell \in \mathcal{N}_{n_{l}}\right)}\left(1-s_{n}(\mathbf{x} ; \bm{\Theta},\mathbf{w})\right)^{\mathbf{1}\left(\ell \in \mathcal{N}_{n_{r}}\right)},
\end{equation}
where $\mathbf{1}(\cdot)$ is an indicator function and $\mathcal{N}_{n_{l}}$ and $\mathcal{N}_{n_{r}}$ denote the sets of nodes (including both split nodes and leaf nodes) held by the sub-trees rooted at the left and right children $n_\texttt{l}$ and $n_\texttt{r}$ of node $n$, respectively. Finally, the output of the tree, \emph{i.e.}, the probability that the category label of $\mathbf{x}$ is y is obtained by
\begin{equation}
  \mathbb{P}_{\texttt{T}}[y|\mathbf{x}, \mathbf{\Theta}, \mathbf{w}, \bm{\pi}]=\sum_{\ell \in \mathcal{L}} \pi_{\ell y} \mu(\ell|\mathbf{x} ; \bm{\Theta},\mathbf{w}),
\end{equation}
where $\bm{\pi}$ are the distributions hold by all the leaves.
We can observe that the decision function $ s_{n}(\mathbf{x} ; \bm{\Theta},\mathbf{w})$ plays an important role in tree building. It is given by
\begin{equation}
    s_{n}(\mathbf{x} ; \bm{\Theta},\mathbf{w})=\sigma\left(f_{\mathcal{\varphi}(n)}(\mathbf{x} ;\bm{\Theta},\mathbf{w})\right),
\end{equation}
where $\sigma(\cdot)$ is a sigmoid function, $f_m(\cdot;\cdot)$ is the output function of neuron $m$ in the new FC layer, and $\varphi(\cdot): \{n_1,n_2,\ldots,n_N\} \rightarrow \{m_1,m_2,\ldots,m_N\}$ is an function to specify the correspondence between the split nodes and the neurons in the new FC layer, \emph{i.e.}, if split node $n$ corresponds to neuron $m$, then $\varphi(n)=m$. In~\cite{Ref:KontschiederFCB15,Ref:ShenGWZWY21}, $\varphi(\cdot)$ is randomly assigned before tree building, and thus results in tree hierarchies without category semantics and unexplainable decisions during inference. Consequently, the key to building a interpretable differentiable decision tree is to learn the correspondence function $\varphi(\cdot)$ to form a tree hierarchy which implicitly clusters semantically-similar categories, so that the sequential decision functions at split nodes along tree paths can correspond to the attributes shared by semantically-similar categories.

\subsubsection{Top-down Hierarchy Learning}
In this section, we describe how to learn the correspondence function $\varphi(\cdot)$ to form a semantically-meaningful tree hierarchy in a top-down manner. Since our strategy is top-down, we first design a criterion function $\mathbb{Q}(m;o)$ to measure how well neuron $m$ can capture category semantics to perform semantically-plausible splitting at the root node $o\in\mathcal{N}$, then we generalize it to any split node $n\in\mathcal{N}$. Intuitively, for a neuron $m$, if its output feature $f_m(\cdot;\cdot)$ is selected for the decision of the root node $o$, \emph{i.e.}, $\varphi(o)=m$, then we expect it can produce similar predictions for samples from semantically-similar categories, \emph{i.e.}, samples from semantically-similar categories are routed to the same sub-tree. Thus, we can define the criterion based on the consistency between category similarities and category-level statistical routing similarities.

Given two categories $c_i,c_j\in\mathcal{Y}$, the category similarity $\mathbb{S}(c_i,c_j)$ between these two can be measured by the similarity between the pre-trained weights $\bm{\omega}_{c_i},\bm{\omega}_{c_j}$ corresponding to these two categories in the original last FC layer (the layer for $C$-way classification) of the CNN $\mathscr{C}$~\cite{Ref:QiaoLSY18,Ref:QiBL18}:
\begin{equation}
\mathbb{S}(c_i,c_j) = \phi(\bm{\omega}_{c_i},\bm{\omega}_{c_j}),
\end{equation}
where $\phi(\cdot,\cdot)$ is the Cosine similarity measure function.
Then, we define the consistency for the two categories $c_i,c_j$ according to the feature selection $\varphi(o)=m$ by
\begin{equation}\label{eq:consistency}
q_{i,j}(m)=\max(\mathbb{S}(c_i,c_j),0)\prod_{c\in\{c_i,c_j\}}\Big(\mathbb{E}_{\mathbf{x}\in\{\mathbf{x}|y=c\}}\left[\sigma(f_m(\mathbf{x} ;\bm{\Theta},\mathbf{w}))\right]-\mathbb{E}_{\mathbf{x}}\left[\sigma(f_m(\mathbf{x} ;\bm{\Theta},\mathbf{w}))\right]\Big),
\end{equation}
where $\mathbb{E}[\cdot]$ is the mathematical expectation, to compute the statistical routing probability for a set of samples. Note that, we apply a Ramp function to the category similarity $\mathbb{S}(c_i,c_j)$, since negative similarity means $c_i$ and $c_j$ are not semantically-similar, and thus we do not consider them. When the category similarity $\mathbb{S}(c_i,c_j)$ is large, we encourage that the statistical routing probabilities for these two categories, \emph{i.e.}, $\mathbb{E}_{\mathbf{x}\in\{\mathbf{x}|y=c_i\}}\left[\sigma(f_m(\mathbf{x} ;\bm{\Theta},\mathbf{w}))\right]$ and $\mathbb{E}_{\mathbf{x}\in\{\mathbf{x}|y=c_j\}}\left[\sigma(f_m(\mathbf{x} ;\bm{\Theta},\mathbf{w}))\right]$, are either both larger or both less than the averaged statistical routing probabilities for all categories, \emph{i.e.},  $\mathbb{E}_{\mathbf{x}}\left[\sigma(f_m(\mathbf{x} ;\bm{\Theta},\mathbf{w}))\right]$
\footnote{Although the parameter $\mathbf{w}$ of the new FC layer is randomly initialized and has not been optimized yet, the  statistical routing probability computed based on $\mathbf{w}$ can still provide sufficient raw discrimination between categories, since the input CNN features of the new FC layer encodes category semantics. This is also coincident to what is found in~\cite{Ref:Frankle21} - \emph{ random features available at initialization provide sufficient raw material to represent high-accuracy functions for image classification.} We will give an evidence later.}.
Otherwise, $q_{i,j}(m)$ becomes negative, which indicates that the output feature $f_m(\cdot;\cdot)$ of neuron $m$ is unable to cluster these two semantically-similar categories into the same sub-tree, and thus the neuron $m$ should not be selected for the decision at the root node $o$. Finally, the criterion function $\mathbb{Q}(m;o)$ for the root node $o$ is given by
\begin{equation}
\mathbb{Q}(m;o)=\sum_{c_i,c_j\in\mathcal{Y},c_i \neq c_j}q_{i,j}(m).
\end{equation}
The corresponding neuron $\varphi(o)$ for the root node $o$ is determined by
\begin{equation}
\varphi(o) = \arg\max_{m\in\mathcal{M}}\mathbb{Q}(m;o).
\end{equation}

Now, we generalize the criterion function $\mathbb{Q}(m;n)$ for any split node $n$. Since the tree hierarchy performs soft data space splitting, the category distribution at a split node $n$ at a deep level of the tree hierarchy is no longer uniform, \emph{i.e.}, the samples of some categories are routed into $n$ with higher probabilities than those of others. Hence, these categories with higher probabilities should be taken into account more in the criterion function $\mathbb{Q}(m;n)$. Towards this end, we calculate a category significance distribution $\bm{\lambda}_n=(\lambda^{(1)}_n,\lambda^{(2)}_n,\ldots,\lambda^{(C)}_n)$ to represent the significance of each category at each split node $n$. This calculation is performed recursively from top to down: At the root node $o$, $\bm{\lambda}_o$ is uniform, \emph{i.e.}, $\lambda^{(c)}_o=\frac{1}{C}$. Then, at its left child $o_\texttt{l}$ and the right child $o_\texttt{r}$,
the category significance distributions $\bm{\lambda}_{o_\texttt{l}}$ and $\bm{\lambda}_{o_\texttt{r}}$ are calculated by
\begin{equation}\label{eq:significance}
{\lambda}_{o_\texttt{l}}^{(c)}=\frac{1}{Z_{o_\texttt{l}}} \gamma^{\beta_o^{(c)}}\lambda^{(c)}_o, {\lambda}_{o_\texttt{r}}^{(c)}=\frac{1}{Z_{o_\texttt{r}}} \gamma^{-\beta_o^{(c)}}\lambda^{(c)}_o,
\end{equation}
where $\beta_o^{(c)}=\mathbb{E}_{\mathbf{x}\in\{\mathbf{x}|y=c\}}\left[\sigma(f_{\varphi(o)}(\mathbf{x} ;\bm{\Theta},\mathbf{w}))\right]-\mathbb{E}_{\mathbf{x}}\left[\sigma(f_{\varphi(o)}(\mathbf{x} ;\bm{\Theta},\mathbf{w}))\right]$ is the bias between the statistical routing probability for category $c$ and the averaged statistical routing probabilities for all categories, $\gamma$ is a control parameter, and $Z_{o_\texttt{l}}$ and $Z_{o_\texttt{r}}$ are normalization factors, to ensure the summation of a category significance distribution is 1. Here we use control parameter $\gamma$ powered with the bias $\beta_o^{(c)}$ to adjust the significance of different categories. In this way, the categories with routing probabilities significantly deviating from the average prediction can obtain higher attention. Then, the categories with routing probabilities close to the average are paid less attention to during feature selection at $o_\texttt{l}$ and $o_\texttt{r}$. This help prevent some categories being routing to an improper path at the beginning.

The category significance distribution $\bm{\lambda}_n$ of each split node $n$ can be obtained by calculating Eq.~\ref{eq:significance} recursively. Then, we can adjust the criterion function $\mathbb{Q}(m;n)$ for a split node $n$ by considering the category significance distribution at the split node $n$:
\begin{equation}
\mathbb{Q}(m;n)=\sum_{c_i,c_j\in\mathcal{Y},c_i \neq c_j}\lambda^{(c_i)}_n\lambda^{(c_j)}_nq_{i,j}(m).
\end{equation}
The corresponding neuron $\varphi(n)$ for each split node $n\in\mathcal{N}$ is determined based on this criterion function $\mathbb{Q}(m;n)$ recursively from top to down. The algorithm of this top-down tree hierarchy learning process is given in Algorithm~\ref{alg:construction}.
\begin{algorithm}[H]
\caption{Top Down Tree Hierarchy Learning}
\label{alg:construction}
\begin{algorithmic}[1]
\State \textbf{Input}: The set of neurons $\mathcal{M}$ in the new FC layer, the set of split nodes $\mathcal{N}$ in the tree
\State \textbf{Output}: The correspondence $\varphi(\cdot)$ between $\mathcal{N}$ and $\mathcal{M}$
\State{Set an empty queue $\mathcal{Q}$, $\mathcal{Q}$.\texttt{push}($o$).} \algorithmiccomment{$o\in\mathcal{N}$ is the root node.}

\While { $\mathcal{Q}$ not empty}
    \State $n$=$\mathcal{Q}$.\texttt{pop}(), $\varphi(n)=\arg\max_{m \in \mathcal{M}} \mathbb{Q}(m;n)$, $\mathcal{M}=\mathcal{M} \backslash \{m\}$.

    \If{$n$ has left and right children $n_\texttt{l}$ and $n_\texttt{r}$}
        \State Calculate $\bm{\lambda}_{n_\texttt{l}}$ and $\bm{\lambda}_{n_\texttt{r}}$ by Eq.~\ref{eq:significance}, $\mathcal{Q}$.\texttt{push}($n_\texttt{l}$), $\mathcal{Q}$.\texttt{push}($n_\texttt{r}$).
    \EndIf
\EndWhile
\end{algorithmic}
\end{algorithm}


\subsubsection{Raw discrimination Provided by Statistical Routing Probabilities}
As we described above, although the parameter $\mathbf{w}$ of the new FC layer is randomly initialized and has not been optimized yet, the statistical routing probability obtained based on $\mathbf{w}$ can still provide raw discrimination between categories~\cite{Ref:Frankle21}. This is because the well pre-trained CNN provides linearly-separable features, so that images from similar categories are distributed closely in high dimensional feature space.

For a randomly initialized $\mathbf{w}$, it is a random hyper-plane in feature space. Each hyper-plane divides categories into left and right sides. The categories distributed closely are more likely to be divided into the same side by a random hyper-plane, so randomly initialized $\mathbf{w}$ can also extract the semantic affinity between categories. The criterion function $\mathbb{Q}(\cdot;\cdot)$ for a random hyper-plane (that is, a neuron), which essentially selects an optimal hyper-plane to split categories to both sides according to the semantic affinity, thus meet the requirements of interpretability. Through subsequent optimization, the hyper-plane is gradually adjusted to increase its confidence in splitting.

\begin{figure}[!htp]

 \centering
 \includegraphics[height=3.5cm]{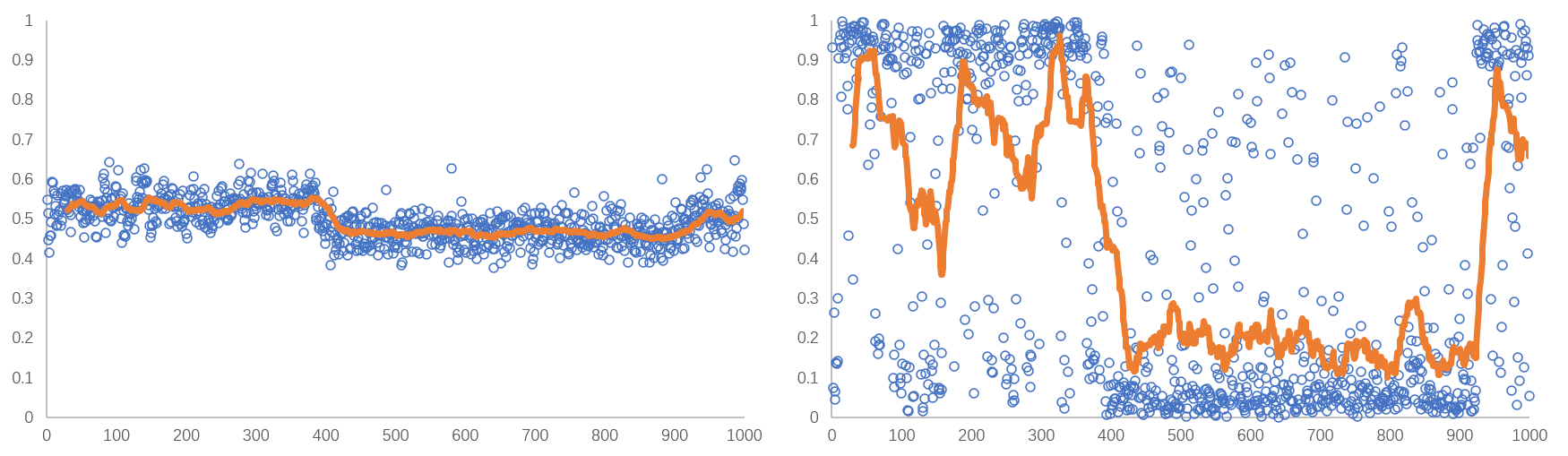}
 \caption{Statistical routing probabilities over categories computed by a randomly initialized $\mathbf{w}$ (left) and a optimized $\mathbf{w}$ (right), respectively.}
 \label{fig:categoryP}

\end{figure}

To demonstrate this, we visualize the statistical routing probability of each category given by the root node $o$ of a tree in our dDSDF built on ImageNet in Fig.~\ref{fig:categoryP}. Fig.~\ref{fig:categoryP} left and right show the statistical routing probabilities over categories computed by a randomly initialized $\mathbf{w}$ and a optimized $\mathbf{w}$, respectively. In Fig.~\ref{fig:categoryP}, each blue point $\left(i,P_{c_i}\right)$, where $P_{c_i}=\mathbb{E}_{\mathbf{x}\in\{\mathbf{x}|y=c_i\}}\left[\sigma(f_o(\mathbf{x} ;\bm{\Theta},\mathbf{w}))\right]$,  represents the statistical routing probability of category $c_i$ given by the root node $o$.

In ImageNet, categories with adjacent indexes often have similar semantics. For example, the first 398 categories are animals, the last 60 categories are plants, and the rest categories are artifacts. Since a good neuron should be able to assign similar statistical routing probabilities to similar categories, the blue points on the plane should distributed like a 1D blob structure, where the boundaries occurs at the significant change between category semantics, \emph{i.e.}, animals $\rightarrow$ artifacts and artifacts $\rightarrow$ plants.
We visualize this 1D blob structure by a orange curve, which is calculated by averaging the statistical routing probabilities of 20 categories with adjacent indexes. It can be observed that the root node can split natural creations and artifacts well even with a randomly initialized $\mathbf{w}$, \emph{i.e.}, the significant changes along the curve only between animals and artifacts and between artifacts and plants. After optimization, the confidence of the statistical routing probabilities become higher, closing to $1$ or $0$. The majority of the categories maintained their pre-optimization routing probabilities, except for a small number of dogs that are split into the artifact side because they often appeared together with the artifacts. This evidences that the discrimination of a randomly initialized neuron can be maintained and improved by optimization.

\subsection{Deep Dynamic Sequential Decision Forest}
A deep Dynamic Sequential Decision Forest (dDSDF) $\mathscr{I}$ is an ensemble of interpretable differentiable decision trees (introduced in Sec.~\ref{sec:tree_building}) built on top of the CNN $\mathscr{C}$ with a dynamic tree selection module (TSM). The dynamic tree selection module is realized by a multi-layer perceptron (MLP) followed by a Softmax layer with $T$ output units, where $T$ is the number of the decision trees in the dDSDF. Thus, each output unit of the TSM provide the probability to select one tree in the forest.

During training, the output of the dDSDF, \emph{i.e.}, the probability that the category label of $\mathbf{x}$ is $y$ is obtained by a weighted average of tree outputs:
\begin{equation}
\mathbb{P}_{\texttt{F}}[y|\mathbf{x},\bm{\Theta},\mathbf{W},\bm{\theta},\bm{\Pi}]= \sum_{t=1}^{T} \alpha_t\mathbb{P}_{\texttt{T}}[y|\mathbf{x},\bm{\Theta},\mathbf{w}^{(t)},\bm{\pi}^{(t)}],
\label{trainEq}
\end{equation}
where $\alpha_t$ (s.t. $\sum_{t=1}^T\alpha_t=1$) is the probability to select $t$-th tree provided by the TSM, $\bm{\theta}$ is the parameter of the TSM, $\mathbb{P}_{\texttt{T}}[y|\mathbf{x},\bm{\Theta},\mathbf{w}^{(t)},\bm{\pi}^{(t)}]$ is the output of the $t$-th tree, $\mathbf{W}=(\mathbf{w}^{(t)};t=1,\ldots,T)$ and $\bm{\Pi}=(\bm{\pi}^{(t)};t=1,\ldots,T)$.
The loss function for the dDSDF is defined as the negative log-likelihood:
\begin{equation}
\mathbb{L}(\bm{\Theta},\mathbf{W},\bm{\theta},\bm{\Pi};\mathbf{x},y)=-\log\big(\mathbb{P}_{\texttt{F}}[y|\mathbf{x},\bm{\Theta},\mathbf{W},\bm{\theta},\bm{\Pi}]\big).
\end{equation}
All the parameters, $\bm{\Theta},\mathbf{W},\bm{\theta},\bm{\Pi}$, are jointly optimized by minimizing the above loss function. Following~\cite{Ref:Shen17,Ref:Shen18,Ref:ShenGWZWY21}, we adopt an alternated optimization strategy: Fix $\bm{\Pi}$, optimize $\bm{\Theta},\mathbf{W},\bm{\theta}$ by Stochastic Gradient Descent; Fix $\bm{\Theta},\mathbf{W},\bm{\theta}$, optimize $\bm{\Pi}$ by Variational Bounding~\cite{Ref:Jordan99,Ref:Yuille03}.

During inference, we only select one tree with the highest probability given by the DTSM, \emph{i.e.}, $t\ast=\arg\max_t\alpha_t$, then the output of the dDSDF is
\begin{equation}
\mathbb{P}_{\texttt{F}}[y|\mathbf{x},\bm{\Theta},\mathbf{W},\bm{\theta},\bm{\Pi}]=\mathbb{P}_{\texttt{T}}[y|\mathbf{x},\bm{\Theta},\mathbf{w}^{(t\ast)},\bm{\pi}^{(t\ast)}].
  \label{valEq}
\end{equation}

\section{Experimental Results}

We verify our method in terms of classification performance and interpretability on several benchmark datasets, including both small-scale datasets, such as Cifar10~\cite{Ref:Krizhevsky09}, Cifar100~\cite{Ref:Krizhevsky09} and TinyImageNet~\cite{Ref:Tiny}, and a large-scale dataset, \emph{i.e.}, ImageNet~\cite{Ref:Russakovsky15}. Since different datasets contain different numbers of categories and image samples, we build dDSDF with different tree numbers ($T$) and depths ($d$) for different datasets: $T=5,d=10$ for Cifar10 and Cifar100, $T=10,d=12$ for TinyImageNet and $T=10,d=14$ for ImageNet. The control parameter $\gamma=10.0$ is set for all the experiments.

\subsection{Implementation Detail}
All the models, including our dDSDF and other competitors, used in the experiments are fine-tuned $200$ epochs based on the pre-trained backbones. The starting learning rate is set as 0.1, and decay by 90\% every 75 epochs. SGD with 0.9 momentum and 5e-4 weight decay is used as the optimizer. We use batch size of 256 on RTX3090 for all experiments.

The dynamic tree selection module (TSM) is a MLP composed of 3 fully connected layers. The first two fully connected layers with ReLU as the activation function  reduce the dimension of input feature to $\frac{1}{4}$, and the third fully connected layer with sigmoid as the activation function outputs the routing probabilites  to $T$ trees.

\begin{table}[!htp]
\centering
\caption{Classification performance on various datasets based on several backbones. R18, R50 and WR28x10 are short for ResNet18, ResNet50 and WideResnet28x10, respectively. We compare our dDSDF with the original neural network (NN) and two deep-decision-tree based methods, deep Neural Decision Forest (dNDF)~\cite{Ref:KontschiederFCB15} and Neural-Backed Decision Tree (NBDT)~\cite{Ref:Wan21}.}
  \label{performance}
  \begin{tabular}{ccccccccc}
  \hline
  \multicolumn{5}{c}{Classification on small-scale datasets}                  &  & \multicolumn{3}{c}{Classification on ImageNet} \\ \cline{1-5} \cline{7-9}
  Model & Backbone        & Cifar10        & Cifar100       & TinyImagenet  &  & Model     & Backbone     & Top1                \\ \cline{1-5} \cline{7-9}
  NN    & R18        & 94.95          & 75.92          & 64.13          &  & NN        & R50     & 76.13               \\
  dNDF  & R18        & 94.96          & 76.02          & 63.96          &  & dNDF      & R50     & 76.298              \\
  NBDT  & R18        & 94.76          & 74.92          & 62.74          &  & dDSDF      & R50     & \textbf{76.49}      \\ \cline{7-9}
  dDSDF  & R18        & \textbf{95.21} & \textbf{76.37} & \textbf{64.20} &  & Model     & Backbone     & Top5                \\ \cline{1-5} \cline{7-9}
  NN    & WR28x10 & 97.62          & 82.09          & 67.65          &  & NN        & R50     & 92.862              \\
  NBDT  & WR28x10 & 97.57          & 82.87          & 66.66          &  & dNDF      & R50     & 92.77               \\
  dDSDF  & WR28x10 & \textbf{97.87} & \textbf{83.11} & \textbf{67.76} &  & dDSDF      & R50     & \textbf{93.316}     \\ \hline
  \end{tabular}

  \end{table}


\subsection{Classification}

We report classification results based on various CNN backbones, including ResNet18~\cite{Ref:HeZRS16}, ResNet50~\cite{Ref:HeZRS16} and WideResNet28x10~\cite{Ref:ZagoruykoK16}. As the comparison shown in Table.~\ref{performance}, dDSDF outperforms the original neural network (NN), based on all backbones. We also compare dDSDF with state-of-the-art deep-decision-tree based models, including dNDF~\cite{Ref:KontschiederFCB15} and NBDT~\cite{Ref:Wan21}. dNDF is a forest-based model, which scarifies interpretability, while NDBT is a tree-based model, which retains interpretable properties, such as sequential decision, non-ensemble.  We set the same tree number and tree depth as dDSDF for dNDF. The tree depth of NDBT is automatically determined by the number of categories. dDSDF achieves better classification performances than both of these two models based on all backbones, and it also enjoys interpretable properties as NDBT.

\subsection{Interpretability}
\subsubsection{Qualitative Interpretability - Sequential Decision Explanation} \label{sec:Qual-Inter}
\begin{figure}[!htp]
\begin{minipage}{0.6\textwidth}
  \centering

  \includegraphics[height=4.5cm]{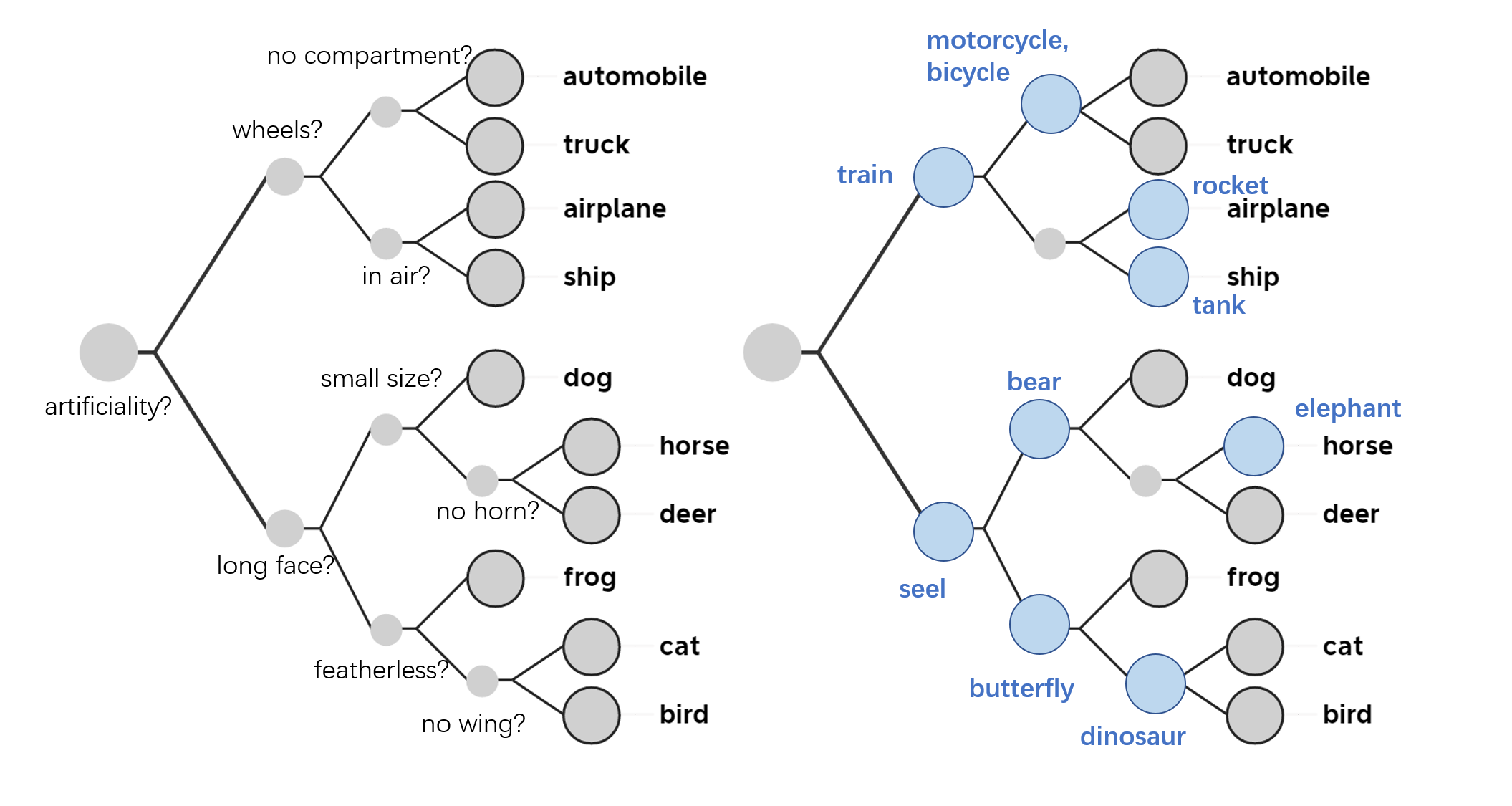}
  \caption{A Cifar10 hierarchy constructed by dDSDF.}
  \label{fig:cifarinter}
\end{minipage}\hfill
\begin{minipage}{0.38\textwidth}
  \centering

  \includegraphics[height=3.5cm]{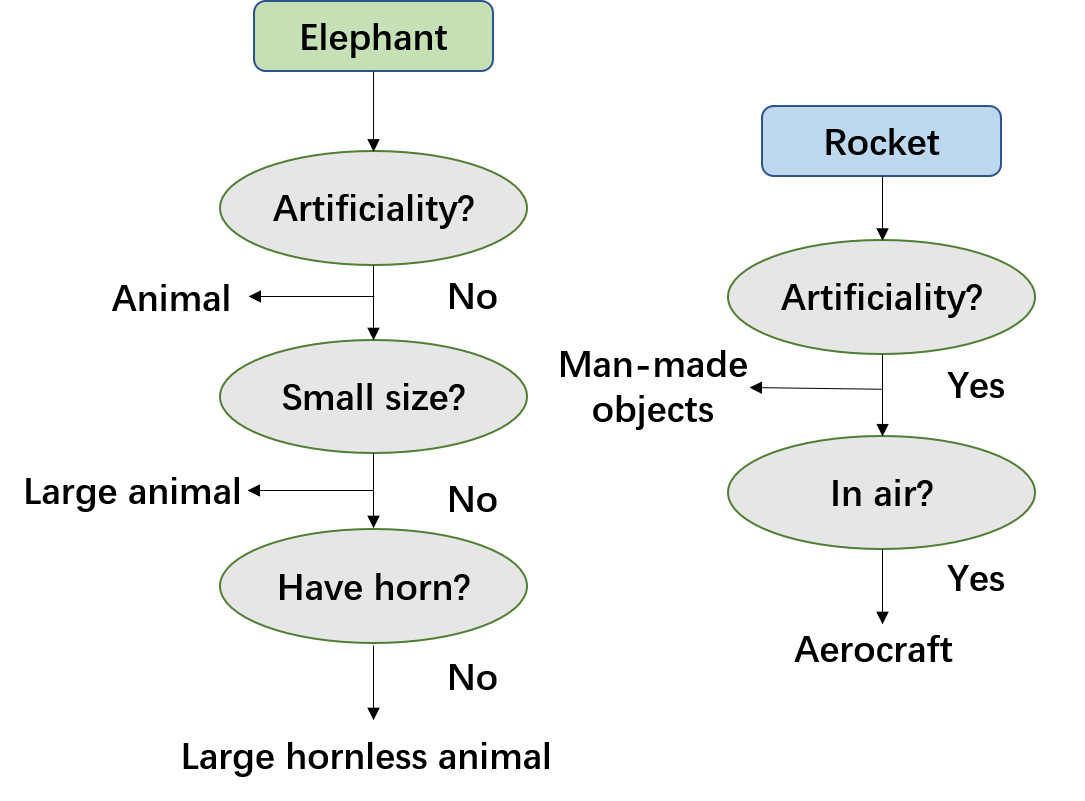}
  \caption{The sequential decision processes provided by dDSDF for OOD samples in Cifar100.}
  \label{fig:path}
\end{minipage}
\end{figure}

To show the interpretability of dDSDF, we investigate whether it can provide interpretable sequential decisions during inference. Since trees in dDSDF perform soft decisions, we define a deterministic decision path during inference for each category $c$ to interpret decisions. Let $\mathcal{P}(n)$ denote the tree path from the root node to split node $n$, then the deterministic decision path for category $c$ is $\mathcal{P}(l^{(c)})$, where $l^{(c)}=\arg\max_{\ell \in \mathcal{L}}\bar{\mu}(\ell|c ; \bm{\Theta},\mathbf{w})$, and $\bar{\mu}(\ell|c ; \bm{\Theta},\mathbf{w})=\prod_{n \in \mathcal{N}} (\bar{s}_n^{(c)})^{\mathbf{1}\left(\ell \in \mathcal{N}_{n_{l}}\right)}\left(1-\bar{s}_n^{(c)})\right)^{\mathbf{1}\left(\ell \in \mathcal{N}_{n_{r}}\right)}$. $\bar{s}_n^{(c)}$ is the  statistical routing probability at split node $n$ for samples from category $c$: $\bar{s}_n^{(c)}= \mathbb{E}_{\mathbf{x}\in\{\mathbf{x}|y=c\}}\left[s_{n}(\mathbf{x};\bm{\Theta},\mathbf{w})\right]$. For two semantically-similar categories $c_i,c_j$, since they are clustered in the same sub-tree in a tree of a trained dDSDF, their deterministic decision paths $\mathcal{P}(l^{(c_i)})$ and $\mathcal{P}(l^{(c_j)})$ share some common split nodes. These split nodes should correspond to some common attributes of the two categories $c_i,c_j$, leading to this sequential decision process. We can specify a semantically-plausible attribute to each shared split node by taking the method proposed in NBDT~\cite{Ref:Wan21}, which makes hypothesise for attributes and verify them by out-of-distribution (OOD) samples. Following NBDT~\cite{Ref:Wan21}, we train a dDSDF with tree number ($T=1$) and depth number ($d=4$) on Cifar10 and take 10 OOD categories from Cifar100. The tree hierarchy and the semantically-plausible attribute of each node are shown in Fig.~\ref{fig:cifarinter} (left). The deterministic decision path $\mathcal{P}(l^{(c)})$ for each OOD category $c$ is shown in Fig.~\ref{fig:cifarinter} (right). Note that, we truncate some deterministic decision paths at split nodes, if the statistical routing probability is close to $0.5$. Specifically, for a deterministic decision path $\mathcal{P}(l^{(c)})$, check each split node along it from the root node, if split node $e$ is the first one which satisfies $|\bar{s}_{e}^{(c)}-0.5|<\tau$, then define $e$ as the end-decision node of this deterministic decision path $\mathcal{P}(l^{(c)})$ and truncate it to $\mathcal{P}(e)$. Based on this hierarchy, we interpret the decision of our model for an input sample by a sequential decision process. Fig.\ref{fig:path} shows the sequential decision processes for an elephant and a rocket given by our method.

\subsubsection{Quantitative Interpretability - Weakly Supervised Localization}
The weakly supervised localization task, \emph{i.e.}, localizing objects in images using image-level category labels only, is commonly used to explain model predictions. The Class Activation Mapping (CAM) approaches~\cite{Ref:ZhouKLOT16,Ref:SelvarajuCDVPB20} are the \emph{de facto} to perform this task, which generate saliency maps by identifying which pixels mostly affected a model's prediction. Here, we propose a decision-tree-based CAM approach, as only one tree is selected in the dDSDF for an input sample during inference.

\paragraph{Decision-tree-based CAM}
First, we define the deterministic decision path $\mathcal{P}(\ell^{(\mathbf{x})})$ for sample $\mathbf{x}$, where $\ell^{(\mathbf{x})}=\arg\max_{\ell \in \mathcal{L}}\mu(\ell|\mathbf{x} ; \bm{\Theta},\mathbf{w})$.
Similarly, we define the end-decision node $e$ for this define the deterministic decision path based on the criterion $ | s_{e}(\mathbf{x};\bm{\Theta},\mathbf{w}) - 0.5|<\tau$. Then use Grad-CAM~\cite{Ref:SelvarajuCDVPB20} to compute the saliency map in terms of end-decision node $e$ as the saliency map $H$ generated by our dDSDF:
\begin{equation}
g_{k}=\frac{1}{Z} \sum_{i} \sum_{j} \frac{\partial \mu(e |\mathbf{x} ; \bm{\Theta},\mathbf{w})}{\partial A_{i j}^{k}},H=\operatorname{ReLU} \left(\sum_{k} g_{k} A^{k}\right),
\label{Eq13}
\end{equation}
where $A_{i j}^{k}$ is the activation value at location $(i,j)$ on the $k$-th feature map of the last convolution layer of the CNN and $Z$ is the normalization factor.

\begin{figure}
\begin{minipage}[b]{0.6\linewidth}

    \centering
    \includegraphics[width=8cm]{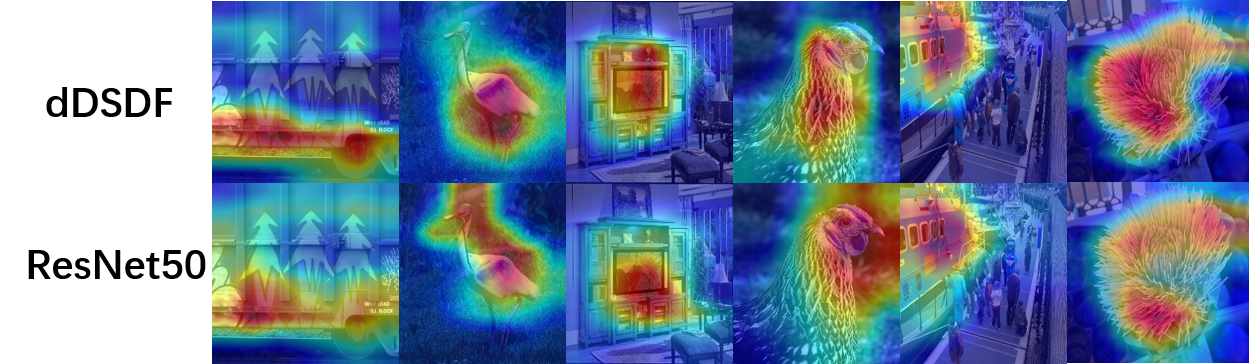}
    \caption{Saliency maps of dDSDT and ResNet50.}
    \label{fig:cam}
\end{minipage}
\begin{minipage}[b]{0.4\linewidth}
\begin{table}[H]
\centering
\caption{Weakly supervised localization results on ImageNet.}
\label{tab:local}
\begin{tabular}{@{}ccc@{}}
\toprule
model    & top1                             & top5                             \\ \midrule
ResNet50 & 38.388                           & 47.098                           \\
dNDF     & 38.396                           & 47.834                           \\
dDSDF     & \textbf{40.236} & \textbf{50.506} \\ \bottomrule
\end{tabular}

\end{table}

\end{minipage}
\end{figure}

\paragraph{Localization results}
In this experiment, we set $\tau=0.1$ and used the same experimental setting as the classification experiment to conduct weakly supervised localization on ImageNet. We use the same setup as Grad-CAM, which sets $15 \% $ of the max intensity as the threshold to get a bounding box on $H$. We report top-1 and top-5 localization accuracies in Table~\ref{tab:local} and compared our result with ResNet50 and dNDF. The saliency maps for ResNet50 and dNDF are generated by their category prediction $y^{c}$, \emph{i.e.}, $g_{k}^{c}=\frac{1}{Z} \sum_{i} \sum_{j} \frac{\partial y^{c}}{\partial A_{i j}^{k}}$. Fig.~\ref{fig:cam} shows the saliency maps obtained by dDSDF and ResNet50, and it shows dDSDF delivers more precise saliency maps.

\subsection{Hyper-partemter Analysis}

In this section, we analyze the hyper-parameters involved in our approach, to see how performance changes by varying them, including tree number ($T$), tree depth ($d$), the control parameter ($\gamma$) to compute category significance distributions, and the threshold ($\tau$) to select the end-decision node for CAM computation. We also discuss the importance of the category similarity $\mathbb{S(\cdot,\cdot)}$. All experiments for hyper-parameter analysis use ResNet18 as the CNN backbone and are conducted on Cifar100, expect for $\tau$, which is discussed on ImageNet using ResNet50. Following the same experimental setting used for classification and localization, we use a dDSDF with $T=5$ and $d=10$ on Cifar100 and a dDSDF with $T=10$ and $d=14$ on ImageNet.

\begin{figure}[!htp]
    \centering
    \includegraphics[width=14cm]{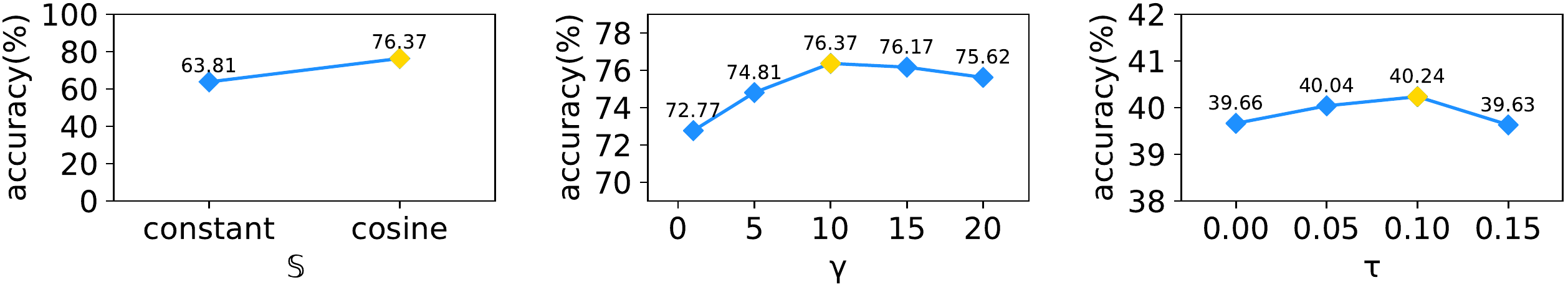}
    \caption{Performance change of dDSDF by varying hyper-parameters on Cifar100.}
    \label{fig:ablation}
\end{figure}

\paragraph{Tree Number and Tree Depth}
We evaluate the performance change by varying tree number ($T$) and tree depth ($d$). The results are shown in Table.~\ref{tab:layertree}. It can be observed that the performance of dDSDF improves significantly with the increase of tree depth. When the tree depth is small, ensemble of more trees can significantly improve the performance, while when the tree depth is large, the benefit of increasing the tree number is weakened.
\begin{table}[!htp]
  \centering
  \caption{Performance change of dDSDF by varying tree number ($T$) and tree depth ($d$) on Cifar100.}
\label{tab:layertree}
\begin{tabular}{c|cccc}
\hline
$d$       & $T=10$ & $T=5$ & $T=3$ & $T=1$ \\ \hline
10  & 76.41    & 76.37   & 76.13   & 75.94   \\
8   & 63.51    & 63.27   & 62.99   & 62.68   \\
6  & 50.83    & 50.67   & 50.14   & 49.74   \\
4  & 17.69    & 16.84   & 15.37   & 14.42   \\ \hline
\end{tabular}

\end{table}
\paragraph{Category Similarity}
To show the importance of the category similarity $\mathbb{S(\cdot,\cdot)}$ in building tree hierarchies, we build a dDSDF by setting $\mathbb{S(\cdot,\cdot)}\equiv1.0$, \emph{i.e,}, tree building does not rely on the category similarity. This leads to a significant performance drop, from 76.37\% to 63.81\% accuracy. In Fig.~\ref{fig:ablation}, ``cosine'' means using the cosine similarity as $\mathbb{S(\cdot,\cdot)}$ and ``constant'' means setting $\mathbb{S(\cdot,\cdot)}\equiv1.0$.

\paragraph{Control Parameter}
The control parameter $\gamma$ adjusts the significance of each category contributing to the criterion function $\mathbb{Q}(\cdot;\cdot)$. To analyze how $\gamma$ influences classification performance, we train and evaluate a dDSDF on Cifar100. The result is shown in Fig.~\ref{fig:ablation}. $\gamma=1$ means the category significance distributions are always uniform, which leads to performance degradation. A very large $\gamma$ forces the criterion function to only take a small number of categories in account, which also leads to a performance drop. Fig.~\ref{fig:ablation} shows the classification accuracy under different $\gamma$.

\paragraph{Threshold for End-decision Node Selection}
Finally, we investigate how the localization performance changes on ImageNet by varying the threshold $\tau$. By decreasing $\tau$, the position of the end-decision node is changed from a shallow level of the tree to a deep level, and thus leads to different sailency maps. We report the localization results with different values of $\tau$ in Fig.~\ref{fig:ablation}, which shows the results are not very sensitive to $\tau$ in a certain range.

\section{Limitation}

 \begin{figure}[htbp]
   \centering
   \includegraphics[width=14cm]{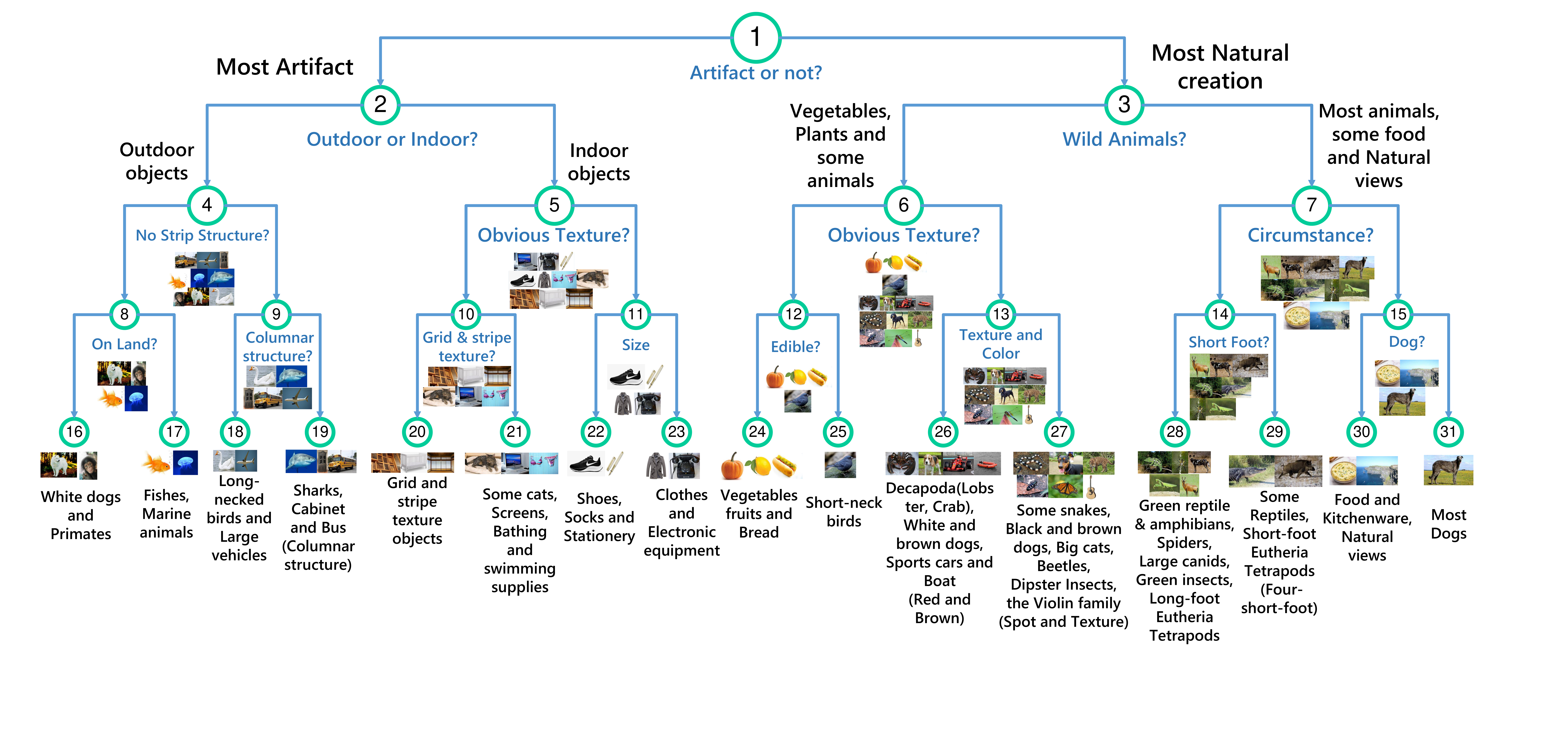} 
   \caption{An ImageNet hierarchy constructed by dDSDF.}
  \label{fig:INT}
 \end{figure}
We build tree hierarchies based on the semantics embedded in the pre-trained weights of the CNN, but such semantics might not consistent with the concepts of categories, especially on large-scale datasets containing a large number of categories. Fig.~\ref{fig:INT} shows the first 5 levels of the hierarchy constructed by dDSDF on ImageNet, according to the deterministic-decision-path based method (Sec.~\ref{sec:Qual-Inter} ). Most of nodes can be specified with a semantically-plausible attribute, like node-$1$, which predicts if an object is an artifact, and node-$3$, which predicts if an object is a wild animal. So it is interesting to validate the semantics of these attributes. We achieve this by generating the saliency map of each corresponding split node by Eq.~\ref{Eq13}.

\begin{figure}[!htp]
    \centering
    \includegraphics[width=12cm]{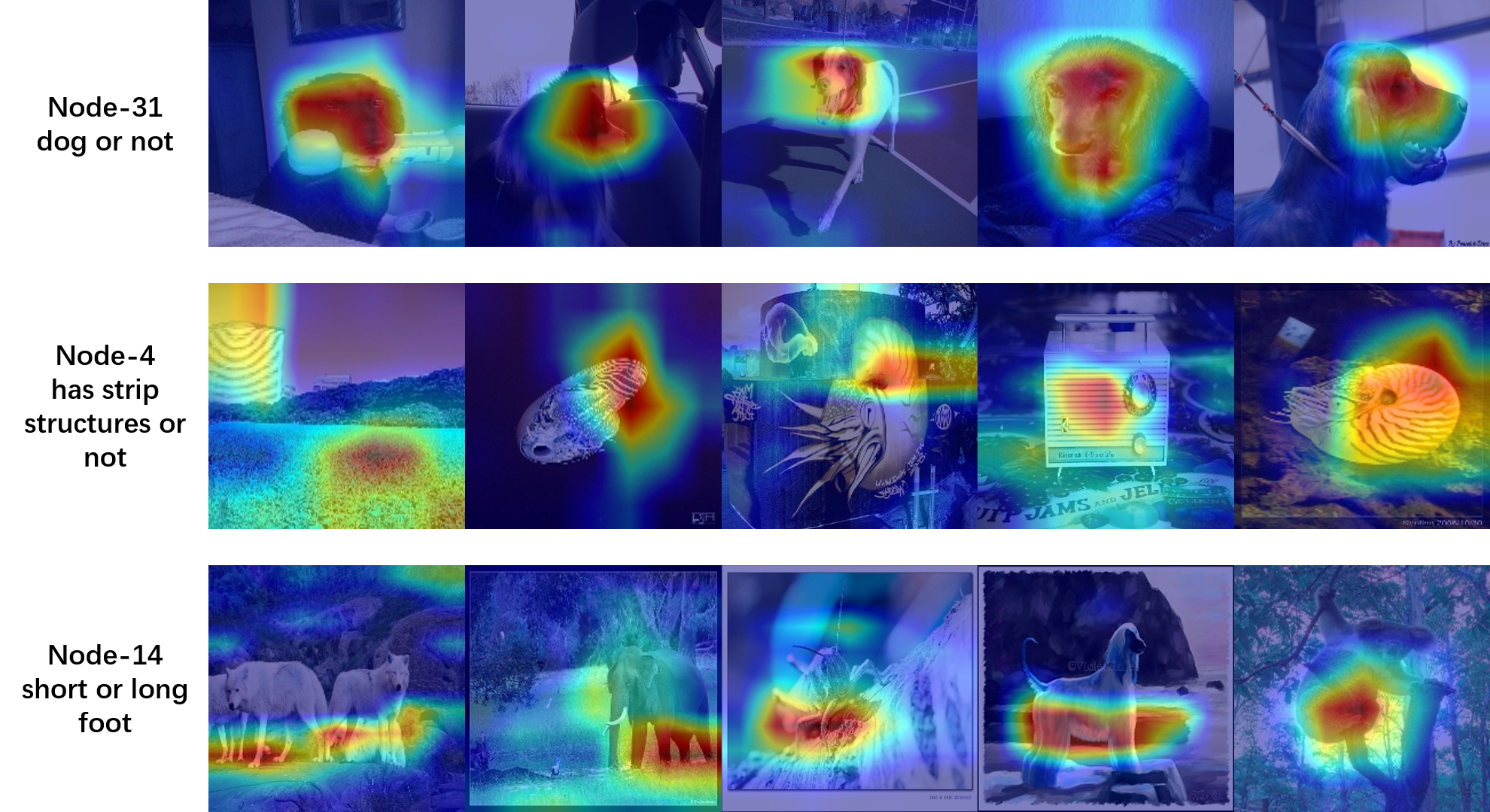}
    \caption{Attribute visualization for split nodes.}
    \label{deep}
\end{figure}

According to Fig.~\ref{fig:INT}, the attributes assigned to node-31, node-4 and node-14 are "dog or not", "has strip structures or not" and "short or long foot", respectively. As shown in Fig.~\ref{deep}, the saliency maps generated at these three split nodes can roughly explain their corresponding attributes. From the top row to the bottom row of Fig.~\ref{deep}, we observe the saliency maps generated at node-31, node-4 and node-14 focus on faces of different varieties of dogs, zebra-like stripes and animal foot.

However, some nodes do not have such a clear attribute consistent with the concepts of categories. For example, cats are branched into the sub-tree of indoor objects by node-$2$ because cats usually show up indoors, and cabinets and buses are grouped together by node-$9$ because they both have columnar structures. There is still large room to improve for building a model with both high performance and good interpretability on large-scale datasets, which is worth exploring in the future.

\section{Conclusion}
We proposed a generic mode transfer scheme to make CNNs interpretable, while maintaining their high classification performance. We achieved this by the proposal of deep Dynamic Sequential Decision Forest. This forest enjoy two properties: 1) Each tree hierarchy in this forest is learned in a top-down manner under the guidance from the category semantics embedded in the pre-trained CNN weights; 2) A dynamical tree selection mechanism is introduced to select one single tree from the forest for each input sample during inference. These two properties enable the forest to make interpreable sequential decisions. Experimental results validated that dDSDF not only achieved higher classification accuracy than the original CNN, but had much better interpretability, both qualitatively and quantitatively.

\bibliographystyle{plain}
\bibliography{papers}

\end{document}